\documentclass[10pt,twocolumn,letterpaper]{article}

\usepackage{cvpr}              

\usepackage{graphicx}
\usepackage{amsmath}
\usepackage{amssymb}
\usepackage{booktabs}
\usepackage{bbding}
\usepackage{color}
\usepackage{pifont}
\usepackage[accsupp]{axessibility}

\usepackage{algorithm, algorithmic}
\renewcommand{\algorithmicrequire}{\textbf{Input:}}  
\renewcommand{\algorithmicensure}{\textbf{Output:}} 

\usepackage[pagebackref,breaklinks,colorlinks]{hyperref}

\usepackage[capitalize]{cleveref}
\crefname{section}{Sec.}{Secs.}
\Crefname{section}{Section}{Sections}
\Crefname{table}{Table}{Tables}
\crefname{table}{Tab.}{Tabs.}

\def\cvprPaperID{1333} 
\def\confName{CVPR}
\def\confYear{2023}

\begin{document}

\title{Towards Accurate Image Coding: \\
Improved Autoregressive Image Generation with Dynamic Vector Quantization}

\author{Mengqi Huang\textsuperscript{\rm 1}, Zhendong Mao\textsuperscript{\rm 1, 2}\thanks{Zhendong Mao is the corresponding author.}, Zhuowei Chen\textsuperscript{\rm 1}, Yongdong Zhang\textsuperscript{\rm 1, 2} \\
\textsuperscript{\rm 1}University of Science and Technology of China,
Hefei, China; \\
\textsuperscript{\rm 2}Institute of Artificial intelligence, Hefei Comprehensive National Science Center, Hefei, China \\
{\tt\small \{huangmq, chenzw01\}@mail.ustc.edu.cn, \{zdmao, zhyd73\}@ustc.edu.cn}
}
\maketitle

\begin{abstract}
   Existing vector quantization (VQ) based autoregressive models follow a two-stage generation paradigm that first learns a codebook to encode images as discrete codes, and then completes generation based on the learned codebook. However, they encode fixed-size image regions into fixed-length codes and ignore their naturally different information densities, which results in insufficiency in important regions and redundancy in unimportant ones, and finally degrades the generation quality and speed. Moreover, the fixed-length coding leads to an unnatural raster-scan autoregressive generation. To address the problem, we propose a novel two-stage framework: (1) Dynamic-Quantization VAE (DQ-VAE) which encodes image regions into variable-length codes based on their information densities for an accurate $\&$ compact code representation. (2) DQ-Transformer which thereby generates images autoregressively from coarse-grained (smooth regions with fewer codes) to fine-grained (details regions with more codes) by modeling the position and content of codes in each granularity alternately, through a novel stacked-transformer architecture and shared-content, non-shared position input layers designs. Comprehensive experiments on various generation tasks validate our superiorities in both effectiveness and efficiency. Code will be released at \url{https://github.com/CrossmodalGroup/DynamicVectorQuantization}.
\end{abstract}

\section{Introduction}
\label{sec:intro}

The vision community has witnessed the rapid progress of deep generative models, pushing image generation quality to an unprecedented level. As a fundamental task, generating realistic images from arbitrary inputs (\eg, class labels) can empower humans to create rich and diverse visual content and bring numerous real-world applications. Unifying the realism of local details and the consistency of global structure is the eternal pursuit for all image generations.

\begin{figure}
  \centering
  \scriptsize
  \includegraphics[width=1.0\linewidth]{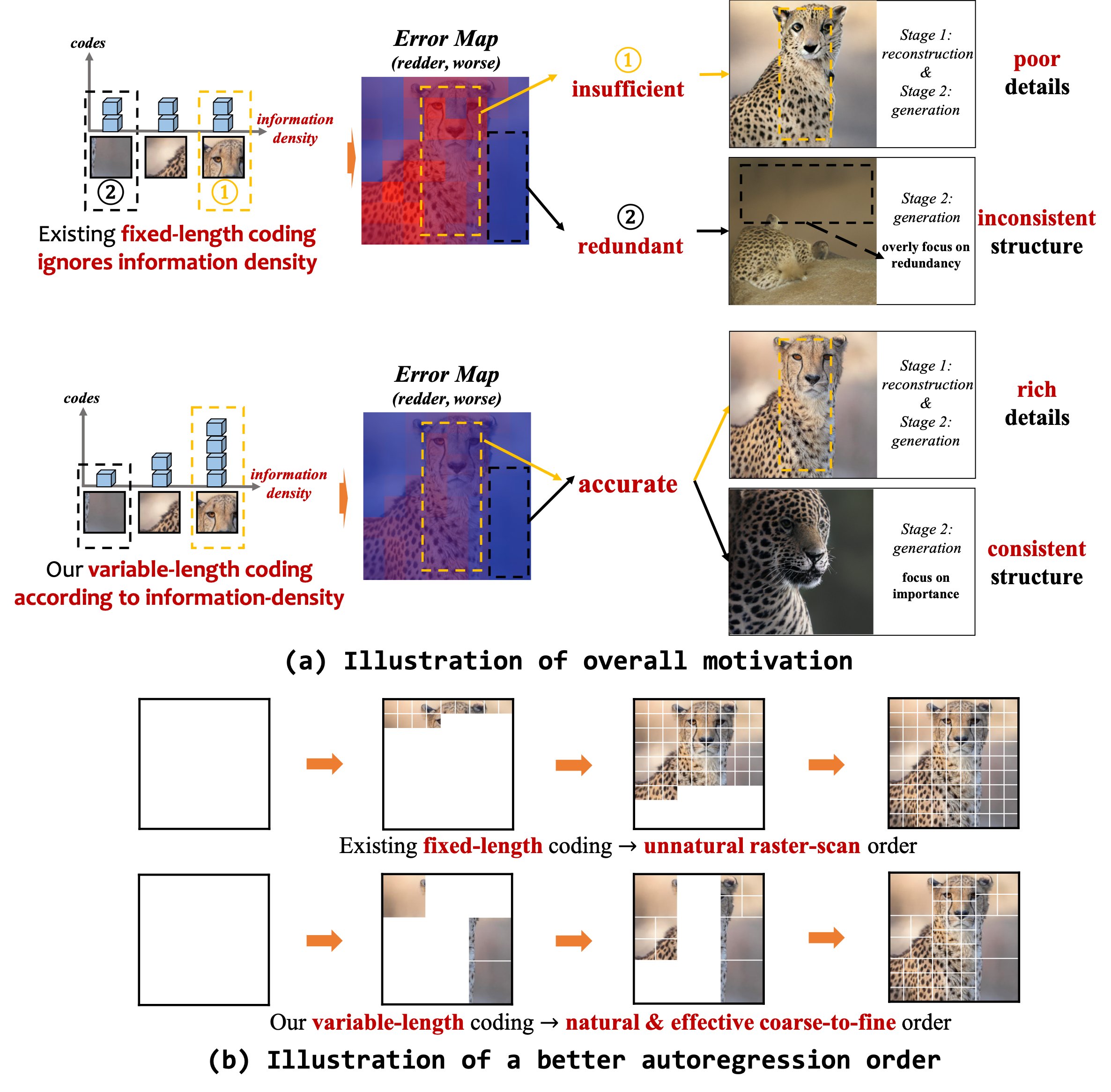}
  \caption{Illustration of our motivation.
  (a) Existing \emph{fixed-length coding} \textbf{ignores information densities}, which results in insufficiency in dense information regions like region \ding{173} and redundancy in sparse information regions like region \ding{172}, generating poor details and inconsistent structure. Our \textbf{information-density-based} \emph{variable-length coding} encodes accurately and produces rich details and consistent structure.
  (b) Comparison of existing unnatural raster-scan autoregressive generation order and our natural and more effective coarse-to-fine autoregressive generation order. 
  \\
  \footnotesize Error map: $l_1$ loss of each $32^2$ region between original images and reconstructions, higher (redder) worse. Existing examples are taken from \cite{esser2021taming}.
  }
  \label{intro}
\end{figure}

Recently, vector quantization (VQ)\cite{van2017neural} has been a foundation for various types of generative models as evidenced by numerous large-scale diffusion models like LDM\cite{rombach2022high}, autoregressive models like DALL-E\cite{ramesh2021zero}, \etc. These models follow a two-stage generation paradigm, \ie, the first stage learns a codebook by VQ to encode images as discrete codes, where each code represents a local visual pattern, while the second stage learns to generate codes of local regions and then restores to images. The importance lies in that the local details could be well encoded in the first stage and thus the second stage could effectively focus on global structure modeling, leading to better generation quality and scalability. Existing models mainly focus on the second stage to better generate codes for improving generation quality, such as raster-scan autoregression\cite{ramesh2021zero, ding2021cogview, yu2022scaling}, bi-direction\cite{chang2022maskgit, lee2022draft, zheng2022movq}, or diffusion\cite{gu2022vector, bond2021unleashing, rombach2022high}. Only \emph{a few} works aim to improve the fundamental code representation itself in the first stage, including perceptual and adversarial loss for context-rich codebook\cite{esser2021taming}, residual quantization\cite{lee2022autoregressive}, and more expressive transformer backbone\cite{yu2021vector}, \etc. Their commonality is that they all focus on encoding more information of all image regions together.

However, existing fundamental encoding works inherently fail to effectively encode image information for an accurate and compact code representation, because they \textbf{ignore the naturally different information densities of different image regions} and encode fixed-size regions into \emph{fixed-length} codes.  As a result, they suffer from two limitations: (1) insufficient coding for important regions with dense information, which fails to encode all necessary information for faithful reconstruction and therefore degrades the realism of local details in both stages.  (2) redundant coding for unimportant ones with sparse information, bringing huge redundant codes that mislead the second stage to focus on the redundancy and therefore significantly hinder the global structure modeling on important ones.  As shown in Figure \ref{intro}(a), the fixed-length codes result in large reconstruction errors in important cheetah regions and produce poor local details (\eg, face, hair) in both stages.  Meanwhile, the fixed-length codes are overwhelmed for unimportant background regions, which misleads the second stage to generate redundant background and inconsistent cheetah structure.  Moreover, as shown in Figure \ref{intro}(b), since all regions are encoded into fixed-length codes, there is no way for the second stage to distinguish their varying importance and thus results in an unnatural raster-scan order\cite{esser2021taming} for existing autoregressive models\cite{ramesh2021zero, lee2022autoregressive, yu2022scaling, ding2021cogview, yu2021vector}, which fails to consider the image content for an effective generation.

To address this problem, inspired by the classical information coding theorems\cite{shannon1948mathematical, shannon1959coding, huffman1952method} and their \textbf{dynamic coding principle}, we propose \textbf{information-density-based} \emph{variable-length coding} for an accurate and compact code representation to improve generation quality and speed. Moreover, we further propose a natural \emph{coarse-to-fine} autoregressive model for a more effective generation. Specifically, we propose a novel two-stage generation framework: (1) \emph{Dynamic-Quantization VAE (DQ-VAE)} which first constructs hierarchical image representations of multiple candidate granularities for each region, and then uses a novel \emph{Dynamic Grained Coding} module to assign the most suitable granularity for each region under the constraint of a proposed \emph{budget loss}, matching the percentage of each granularity to the desired expectation holistically. (2) \emph{DQ-Transformer} which thereby generates images autoregressively from coarse-grained (smooth regions with fewer codes) to fine-grained (details regions with more codes) to more effectively achieve consistent structures. Considering the distribution of different granularities varying, DQ-Transformer models the position and content of codes in each granularity alternately through a novel \emph{stacked-transformer architecture}. To effectively teach the difference between different granularities, we further design \emph{shared-content} and \emph{non-shared-position} input layers.


Our main contributions are summarized as follows:

\textcolor{blue}{\textbf{Conceptual contribution.}} We point to the inherent insufficiency and redundancy in existing \emph{fixed-length coding} since they \textbf{ignore information density}. For the first time, we propose \textbf{information-density-based} \emph{variable-length coding} for accurate $\&$ compact code representations.

\textcolor{blue}{\textbf{Technical contribution.}} (1) We propose \emph{DQ-VAE} to dynamically assign variable-length codes to regions based on their different information densities through a novel \emph{Dynamic Grained Coding module} and \emph{budget loss}. (2) We propose \emph{DQ-Transformer} to generate images autoregressively from coarse-grained to fine-grained for the first time, which models the position and content of codes alternately in each granularity by  \emph{stacked-transformer architecture} with \emph{shared-content} and \emph{non-shared position} input layers design.

\textcolor{blue}{\textbf{Experimental contribution.}} Comprehensive experiments on various generations validate our superiority, \eg, we achieve 7.4\% quality improvement and faster speed compared to existing state-of-the-art autoregressive model on unconditional generation, and 17.3\% quality improvement compared to existing million-level parameters state-of-the-art models on class-conditional generation.

\section{Related Works}

\begin{figure*}
  \centering
  \includegraphics[width=1.0\linewidth]{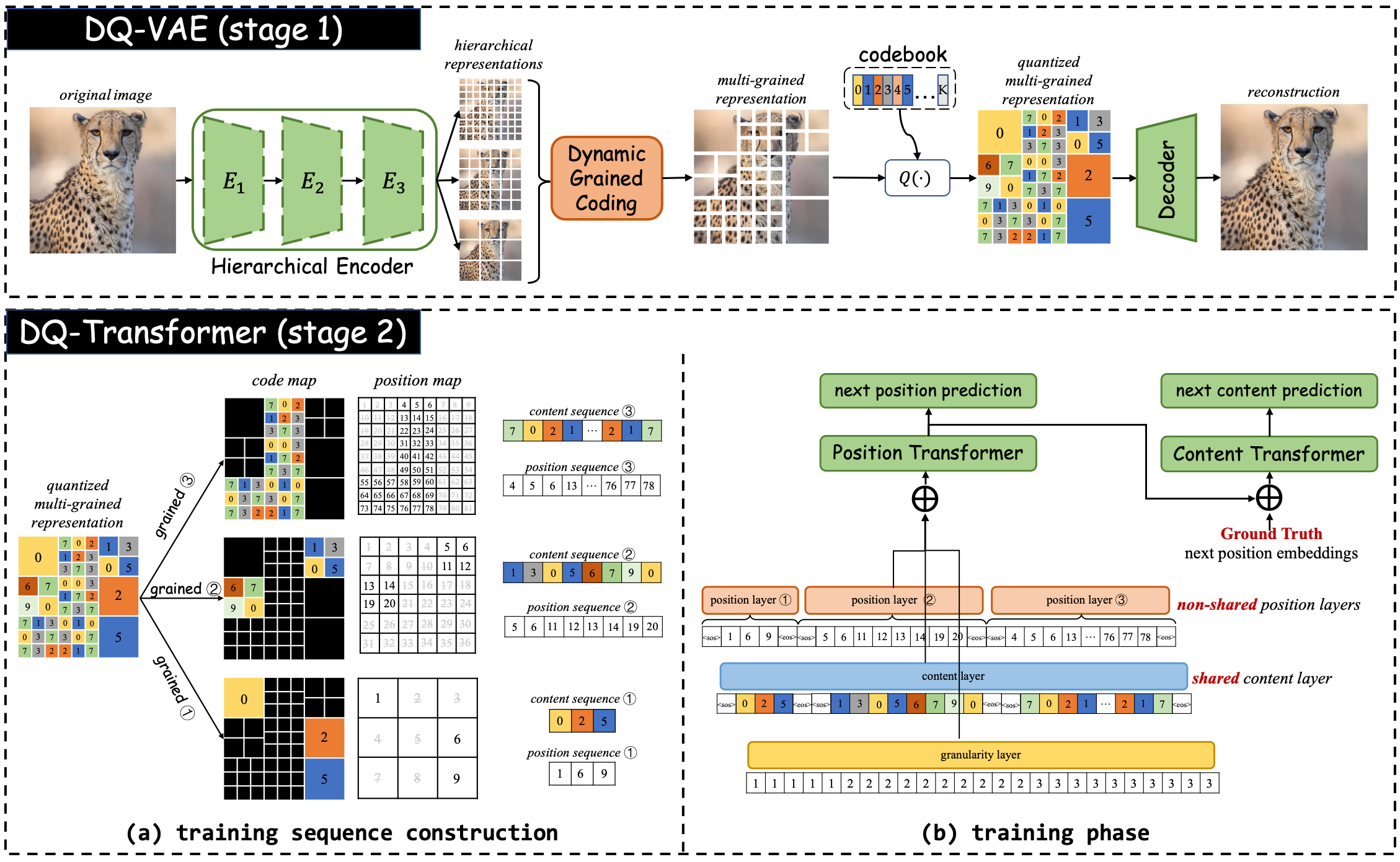}
  \caption{The overview of our proposed two-stage framework. (1) DQ-VAE dynamically assigns variable-length codes for each image region through \emph{Dynamic Grained Coding (DGC)} module. (2) DQ-Transformer models the position and content of codes alternately by the stacked \emph{Position-Transformer} and \emph{Content-Transformer}, generating images autoregressively from coarse-grained to fine-grained. To effectively teach the difference between granularities, we further design \emph{shared-content}, \emph{non-shared-position}, and \emph{granularity} input layers.
  }
  \label{framework}
\end{figure*}

\subsection{Vector Quantization for Image Generation}

Existing \emph{VQ-based} models follow a two-stage paradigm that first learns a codebook to encode images into discrete space and then models the underlying distribution in this discrete space. The \emph{VQ-based} paradigm has attracted increasing interest and is adopted by most milestone generative models, such as latent diffusion\cite{rombach2022high}, DALL-E\cite{ramesh2021zero}, Parti\cite{yu2022scaling}, \etc. Most works focus on the second stage for better learning in the discrete space, such as discrete diffusion\cite{gu2022vector, austin2021structured, tang2022improved, zhu2022discrete, rombach2022high}, bidirection\cite{chang2022maskgit, lee2022draft, zheng2022movq, bond2021unleashing}, and the most popular raster-scan autoregression\cite{ramesh2021zero, ding2021cogview, van2017neural, esser2021taming, lee2022autoregressive, yu2022scaling, gu2022vector, razavi2019generating}. Only a few works aim to improve the fundamental encoding, \eg, VQGAN\cite{esser2021taming} introduces perceptual and adversarial loss for a context-rich codebook. \cite{lee2022autoregressive} introduces residual-quantization. \cite{yu2021vector} proposes a more expressive transformer backbone. Recently, \cite{zheng2022movq} proposes to insert spatially variant information. However, existing fixed-length coding ignores information density and is thus limited by insufficiency and redundancy. For the first time, we propose information-density-based variable-length coding and a more natural coarse-to-fine autoregression.

\subsection{Dynamic Network}

Designing dynamic architectures is an effective approach for efficient deep learning and yields better representation power and generality\cite{han2021dynamic}. Literately, current research can be mainly categorized into three directions, \ie, dynamic depth for network early exiting\cite{bolukbasi2017adaptive} or layer skipping\cite{veit2018convolutional}, dynamic width for skipping neurons\cite{bengio2015conditional} or channels\cite{liu2017learning} and dynamic routing for multi-branch structure networks\cite{huang2022dse, li2020learning, song2021dynamic, yang2020resolution}. Our work belongs in the last direction. To the best of our knowledge, the dynamic network
has never been studied in VQ-based generation and we present the first work to realize the variable-length coding of classical information coding theorems through the dynamic network.

\section{Methodology}

Our overall two-stage framework is depicted in Figure \ref{framework}. In the following, we will first briefly revisit the formulation of VQ and then describe our proposed method in detail.


\subsection{Preliminary}

Vector Quantization (VQ)\cite{van2017neural} denotes the technique that learns a codebook to encode images into discrete code representations.  Formally, the codebook is defined as $\mathcal{C} := \{(k, \boldsymbol{e}(k))\}_{k \in [K]}$, where $K$ is the codebook size and $n_z$ is the dimension of codes. An image $\mathbf{X} \in \mathbb{R}^{H_0 \times W_0 \times 3}$ is first encoded into grid features $\mathbf{Z} = E(\mathbf{X}) \in \mathbb{R}^{H \times W \times n_z}$ by the encoder $E$, where $(H, W) = (H_0/f, W_0/f)$ and $f$ is the downsampling factor. For each vector $\boldsymbol{z} \in \mathbb{R}^{n_z}$ in $\mathbf{Z}$, the quantization operation $\mathcal{Q}(\cdot)$ replaces it with the code that has the closest euclidean distance with it in the codebook $\mathcal{C}$:

\begin{equation}
    \mathcal{Q}(\boldsymbol{z}; \mathcal{C}) = \arg\min\limits_{k \in [K]} || \boldsymbol{z} - \boldsymbol{e}_k ||^{2}_{2}.
\end{equation}
Here, $\mathcal{Q}(\boldsymbol{z}; \mathcal{C})$ is the quantized code. $\boldsymbol{z^q} = \boldsymbol{e}(\mathcal{Q}(\boldsymbol{z}; \mathcal{C}))$ is the quantized vector. Therefore, the quantized encoded features are $\mathbf{Z}^{\boldsymbol{q}} \in \mathbb{R}^{H \times W \times n_z}$. The decoder $D$ is used to reconstruct the original image by $\Tilde{\mathbf{X}} = D(\mathbf{Z}^{\boldsymbol{q}})$. Here each code roughly represents a fixed $f^2$ size visual pattern and each image region is represented by the same length of codes without distinguishing their different information densities. As a result, existing works suffer from both insufficiency in important regions and redundancy in unimportant ones.

\subsection{Stage 1:Dynamic-Quantization VAE(DQ-VAE)}

Different from existing works that adopt a fixed downsampling factor $f$ to represent image regions as fixed-length codes, DQ-VAE first defines a set of candidates:

\begin{equation}
    {\rm F} = \{f_1, f_2, ..., f_K \}, \text{where } f_1 < f_2 < ... < f_K,
\end{equation}
and encodes images into hierarchical features $\mathbf{Z} = \{\mathbf{Z}_1, \mathbf{Z}_2, ..., \mathbf{Z}_K \}$ through a hierarchical encoder $E_h$, where $\mathbf{Z}_i \in \mathbb{R}^{H_i \times W_i \times n_z}$ and $(H_i, W_i) = (H_0 / f_i, W_0 / f_i)$ for each $i \in \{1, 2, ..., K\}$. The image region's size is set as the maximum downsampling factor, \ie, $S=f_{K}$, and therefore each $S^2$ size image region now has multiple granularity representations containing different numbers of features. Then the \emph{Dynamic Grained Coding (DGC)} module assigns the most suitable granularity for each region and results in multi-grained representations, which are further quantized by VQ. To deal with the irregular code map that different regions have different numbers of codes, we further propose a simple but effective nearest-neighbor replication, that is, in each region the quantized codes are replicated to the code number of the finest granularity if the finest granularity is not assigned for it, resulting in a regular code map that could be conveniently decoded by the convolutional decoder $D$. 

\textbf{Dynamic Grained Coding (DGC) module.} As illustrated in Figure \ref{dgc_module}, given the encoded hierarchical image features $\mathbf{Z} = \{\mathbf{Z}_1, \mathbf{Z}_2, ..., \mathbf{Z}_K \}$, we implement a discrete gating network with Gumbel-Softmax technique\cite{jang2016categorical} to determine the granularity for each image region. Specifically, each granularity feature is first normed by group-normalization to stabilize training and then pooled to the size of the coarsest granularity feature by average-pooling, except the coarsest granularity (\ie, $f_K$) feature itself. The pooled features are denoted as $\mathbf{Z^{'}} = \{\mathbf{Z}^{'}_1, \mathbf{Z}^{'}_2, ..., \mathbf{Z}^{'}_K \}$ and $\mathbf{Z}^{'}_i \in \mathbb{R}^{H_s \times W_s \times n_z}$ for $i \in \{1,2,...,K\}$, where $(H_s, W_s) = (H_0 / f_{K}, W_0 / f_{K})$. The gating logits $\mathbf{G}$ are generated as:

\begin{figure}
  \centering
  \includegraphics[width=1.0\linewidth]{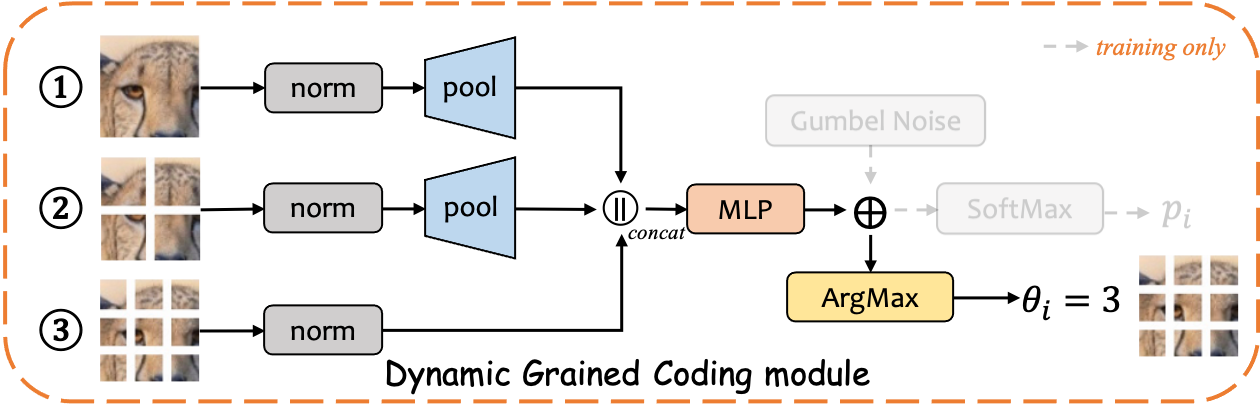}
  \caption{Illustration of our Dynamic Grained Coding module.}
  \label{dgc_module}
\end{figure}

\begin{equation}
    \mathbf{G} = (\mathbf{Z}^{'}_1 \Vert \mathbf{Z}^{'}_2 \Vert ... \Vert \mathbf{Z}^{'}_K)\mathbf{W_g} \in \mathbb{R}^{H_s \times W_s \times k},
\end{equation}
where $\Vert$ is the concatenation operation along the channel dimension and $\mathbf{W_g} \in \mathbb{R}^{(K \times n_z) \times K}$ is the learnable weight. For each region $(i,j)$, its gating logits $g_{i,j} \in \mathbb{R}^{K}$ is used to decide the granularity by calculating the gating index:

\begin{equation}
    \label{gate}
    \theta_{i,j} = \underset {k} { \operatorname {arg\,max} }(g_{i,j,k}) \in \{1,2,...,K \}.
\end{equation}
To enable the end-to-end training of this discrete process, inspired by \cite{xie2020spatially, zhu2021make}, the determined decisions in Eq.(\ref{gate}) are replaced with the stochastic sampling process. Mathematically, given a categorical distribution with unnormalized log probabilities, discrete gating indices can be yielded with noise samples drawn from a standard Gumbel distribution:

\begin{equation}
    \small
    \theta_{i,j} = \underset {k} { \operatorname {arg\,max} }(g_{i,j,k} + n_k), \text{where } n_k  \sim \text{Gumbel(0,1)}.
\end{equation}
To enable the back-propagation of the above hard decision, we adopt the Gumbel-Softmax technique\cite{jang2016categorical} to give a continuous and differentiable approximation by replacing the argmax with a softmax operation. The soft gating score $p_{i,j}$ for each region is then selected by the gating indices:

\begin{equation}
    p_{i,j} = \frac{\exp((g_{i,j,\theta_{i,j}} + n_{\theta_{i,j}})) / \tau}{\sum_k^K \exp((g_{i,j,k} + n_k) / \tau)} \in [0,1],
\end{equation}
where the temperature $\tau=1$. We use a straight-through estimator for the gradients of gating logits, which are obtained through the soft gating score $p_{i,j}$ during the backward pass. The above stochastic process is only adopted during training and no random sampling is required during inference.

\textbf{Budget Loss.} We adopt the training loss of VQGAN\cite{esser2021taming} as $\mathcal{L}_{vanilla}$, which includes reconstruction loss ($l_1$ loss,  perceptual loss, adversarial loss) and quantization loss. In the absence of a budget constraint, the DGC module typically prefers to assign the finest granularity for all image regions, which is in contrast to our purpose. Therefore, we further propose a \emph{budget loss} to match the percentage of each granularity to our desired expectation. Specifically, we denote the desired ratio of each granularity $k$ as $r_k$ and $\sum_k^K r_k = 1$. For an image sample whose current assigned ratio of each granularity $k$ is $r_k^{'}$, we define \emph{budget loss} as:

\begin{equation}
    \mathcal{L}_{budget} = \sum_k^{K-1}(r_k - r_k^{'})^2,
\end{equation}
where we only calculate on $K-1$ granularities since the ratio of the last granularity is determined by $1-\sum_k^{K-1}r_k$. The final loss for DQ-VAE is defined as:

\begin{equation}
    \mathcal{L}_{stage1} = \mathcal{L}_{vanilla} + \lambda \mathcal{L}_{budget},
\end{equation}
where $\lambda$ is a loss balance hyper-parameter. The expected ratio of each granularity is holistic on the dataset level. Therefore, since important regions contribute more to the reconstruction quality, the variable-length coding is realized from two aspects, \ie, \emph{inter-dynamic}, longer code sequence for complex images while shorter code sequence for easy ones; \emph{intra-dynamic}, for each image, more codes for important regions while fewer codes for unimportant ones.

\subsection{Stage 2: DQ-Transformer}
Different images share different perceptually important regions and different complexities. Therefore, DQ-VAE encodes images as the code sequence of variable lengths and the distribution of each granularity region in images is also completely different. Though learning this dynamic underlying prior is very challenging, it also opens a promising potential for autoregressive image generation, that is, a natural and more effective coarse-grained to fine-grained generation order since DQ-VAE naturally divides coarse regions (smooth regions with fewer codes) apart from fine regions (details regions with more codes). Imagine image generation as a jigsaw puzzle problem, it is more effective and efficient that we first fill in the large and easy pieces (coarse regions) and then fill in the small and difficult ones (fine regions). With this motivation, DQ-Transformer first constructs the codes' content and position sequence in each granularity separately and then concatenates them in a coarse-to-fine manner to autoregressively predict the next code's position and content through the stacked \emph{Position-Transformer} and \emph{Content-Transformer}. The distinction of different granularities is realized by the \emph{shared content}, \emph{non-shared-position}, and \emph{granularity} input layers designs.

\textbf{Training sequence construction.} As illustrated in stage 2(a) in Figure \ref{framework}, the sequence of each granularity is constructed separately. As for the content sequence, each index is the quantized code index. As for the position sequence, each index is the position of the corresponding code index in the \emph{position map of current granularity}. We add a special $<$sos$>$ code at the beginning of all content and position sequences to indicate the start of the sequence, and another special $<$eos$>$ code at the end of them to indicate the end of the sequence. To enable batch training and sampling, we use a special $<$pad$>$ code to pad all samples to the same length in each granularity. Finally, we concatenate all granularities' content and position sequences in a coarse-to-fine manner, which we denote as $C$ and $P$, respectively.

\textbf{Position-Transformer.} We first learn to predict the next code position conditioned on all previous codes and their positions. The input of Position-Transformer consists of four parts: (1) content embedding which is calculated from $C$ by a \emph{shared-content} layer for all granularities, (2) position embedding which is calculated from $P$ by \emph{non-shared-position} layers for each granularity separately, (3) granularity embedding which is used for distinguishing each granularity, and (4) a learned absolute position embedding for making the network aware of the absolute position of the sequence, which is the same as most transformer-architecture\cite{vaswani2017attention, tang2022improved, esser2021taming}. After processing by Position-Transformer, the output hidden vector $H_p$ encodes both code and their position information and is used for next position predicting. The negative log-likelihood (NLL) loss for the next code position autoregressive training is:

\begin{equation}
    \mathcal{L}_{position} = \mathbb{E}(-\log p(P_l | P_{< l}, C_{< l}))
\end{equation}

\textbf{Content-Transformer.} We then learn to predict the next code's content conditioned on all previous codes and the position of \emph{current} code. Specifically, The input of Content-Transformer is two parts: (1) the output of Position-Transformer $H_p$ and (2) the ground-truth information of the current position which also is calculated by the \emph{non-shared-position} layers. For example, if the input position sequence for Position-Transformer is $P_{[0:-2]}$, then the input ground-truth position sequence for Content-Transformer is $P_{[1:-1]}$. The negative log-likelihood (NLL) loss for the next code's content autoregressive training is:

\begin{equation}
    \mathcal{L}_{content} = \mathbb{E}(-\log p(C_l | P_{\leq l}, C_{< l}))
\end{equation}

\textbf{Training $\&$ Inference.} During training, the total loss for DQ-Transformer is defined as:

\begin{equation}
    \mathcal{L}_{stage2} = \mathcal{L}_{position} + \mathcal{L}_{content}.
\end{equation}
Our proposed DQ-Transformer is a general visual generative model which could be easily extended to various other generation tasks. As for the class-conditional generation, we replace the $<$sos$>$ code in the content sequence of each granularity with the class-label code. During inference, we could also autoregressively generate images from coarse-grained to fine-grained, as illustrated in Algorithm \ref{alg:sample}, where we take the unconditional generation as an example and other conditional generations can be derived accordingly. 

\begin{algorithm}[!h]
    \small
    \caption{Unconditional batch sampling.}
    \label{alg:sample}
    \renewcommand{\algorithmicrequire}{\textbf{Input:}}
    \renewcommand{\algorithmicensure}{\textbf{Output:}}
    \begin{algorithmic}[1]
        \REQUIRE 
        The granularity number $K$ and batch size $B$; \\
        The initial empty position (code) sequence $P$ ($C$).
      
        \ENSURE The generated image $\mathcal{I}$.    

        \FOR{each $k \in [1,K]$}
            \STATE // sample each granularity in a coarse-to-fine order
            \STATE $P = \operatorname{concat}(P,<$sos$>)$, $C = \operatorname{concat}(C,<$sos$>)$
            \WHILE{NOT all samples have sampled $<$eos$>$}
                \STATE mask sampled position indexes to avoid repeat
                \STATE sample next code position $P_i \in \mathbb{R}^{B}$
                \IF {$P_{i,b} == <$eos$>$, for $b \in [1,B]$}
                    \STATE $P_{>i,b}$ = $<$pad$>$
                    \STATE // if sampled $<$eos$>$, the following will only can be $<$pad$>$ for this sample in current granularity
                \ENDIF
                \STATE sample next code $C_i$
                \STATE $C = \operatorname{concat}(C, C_i)$, $P = \operatorname{concat}(P, P_i)$
            \ENDWHILE
        \ENDFOR
        \RETURN decoded image $\mathcal{I}$ from $P$ and $C$
    \end{algorithmic}
\end{algorithm}

\begin{figure*}
  \centering
  \includegraphics[width=1.0\linewidth]{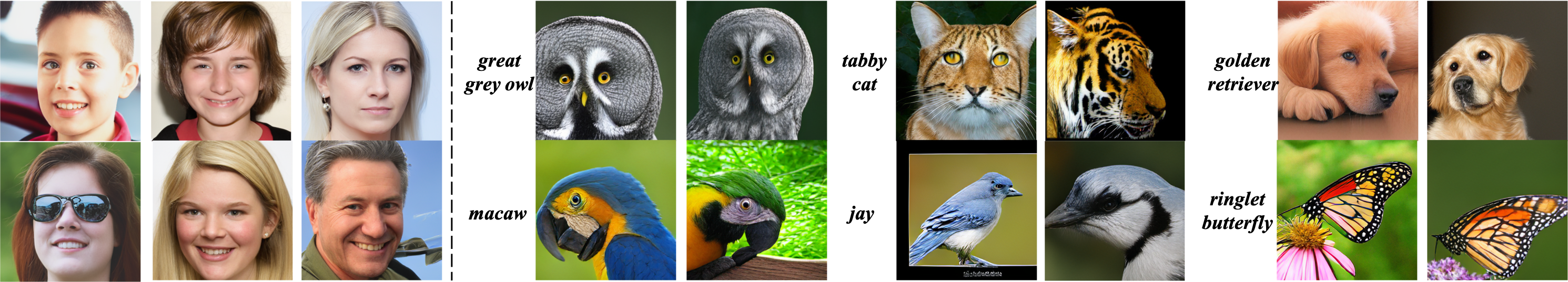}
  \caption{Qualitative results. Left: Our unconditional generation on FFHQ. Right: Our class-conditional generation on ImageNet.}
  \label{visual}
\end{figure*}

\section{Experiments}


\textbf{Benchmarks.} We evaluate our method on unconditional FFHQ\cite{karras2019style} benchmark and class-conditional ImageNet\cite{deng2009imagenet} benchmark with $256 \times 256$ image resolution.

\textbf{Metrics.} Following previous works\cite{esser2021taming, lee2022autoregressive,zheng2022movq}, the standard Fréchet Inception Distance (FID)\cite{heusel2017gans} is adopted for evaluating the generation and reconstruction quality (denoted as rFID). rFID is calculated over the entire test set. Inception Score (IS)\cite{heusel2017gans} is also adopted for class-conditional generation on the ImageNet benchmark. 

\textbf{Implementation.} DQ-VAE follows the architecture of VQGAN\cite{esser2021taming} except for the lightweight DRC module, which is trained with the codebook size $K=1024$ and $\lambda=10$. DQ-Transformer adopts a stack of causal self-attention blocks\cite{vaswani2017attention} and is trained with two different settings, \ie, DQ-Transformer$_b$(base) with $6$ layers Position-Transformer and $18$ layers Content-Transformer of a total 308M parameters, and DQ-Transformer$_l$(large) with $6$ layers Position-Transformer and $42$ layers Content-Transformer of a total 608M parameters to demonstrate our scalability. All models are trained with eight RTX-3090 GPUs. Top-k and top-p sampling are used to report the best performance. More details can be found in the supplementary.

\subsection{Comparison with state-of-the-art methods}

The main results are reported on dual granularities of $\rm F=\{8,16\}$, and the ratio $r_{f=8}=0.5$ (640 average length).

\begin{table}[t]
    \small
	\centering
	\begin{tabular}{l c c c}
		\toprule
		Model & $L$ & \#Params & FID$\downarrow$ \\
		\hline 
		VQGAN$_{('21)}$\cite{esser2021taming} & 256 & (72.1+307)M & 11.4 \\
		DCT$_{('21)}$\cite{nash2021generating} & $>$1024 & 738M & 13.06 \\
		ViT-VQGAN$_{('22)}$\cite{yu2021vector} & 1024 & (599+1697)M & 5.3 \\
		RQ-VAE$_{('22)}$\cite{lee2022autoregressive} & 256 & (100+355)M & 10.38 \\
		Mo-VQGAN$_{('22)})$\cite{zheng2022movq} & 1024 & (82.7+307)M & 8.52 \\
		\hline
		\textbf{DQ-Transformer$_b$} & 640 & \textbf{(47.5+308)}M & \textbf{4.91} \\
		\bottomrule
	\end{tabular}
	\caption{\footnotesize Comparison of unconditional autoregressive generation on FFHQ. $L$ is coding length. \#Params splits in (VAE+autoregressive model).}
	\label{unconditional_autoregression}
\end{table}

\begin{table}[t]
    \footnotesize
	\centering
	\begin{tabular}{l c c c c c c}
		\toprule
		Type & Model & $L$ & \#Params & FID$\downarrow$ & IS$\uparrow$ \\
		\hline
		GAN & \scriptsize BigGAN-deep\cite{brock2018large} & - & 160M & 6.95 & 198.2 \\
		\hline
		diffusion & \cite{nichol2021improved} & - & 280M & 12.26 & - \\
		  diffusion & ADM\cite{dhariwal2021diffusion} & - & 554M & 10.94 & 101.0 \\
		\hline
		  bi-direct & MaskGIT\cite{chang2022maskgit} & 1024 & 227M & 6.18 & 182.1 \\
		\hline
		ARM & VQGAN*\cite{esser2021taming}& 256 & 379M & 17.5 & 75 \\
		ARM & DCT\cite{nash2021generating}& $>$1024 & 738M & 36.5 & - \\
		ARM & RQ-VAE\cite{lee2022autoregressive} & 256 & 480M & 15.72 & 86.8 \\
		ARM & RQ-VAE\cite{lee2022autoregressive} & 256 & 821M & 13.11 & 104.3 \\
		ARM & \scriptsize Mo-VQGAN\cite{zheng2022movq} & 1024 & 389M & 7.12 & 138.3 \\
		\hline
		ARM & \scriptsize DQ-Transformer$_b$ & 640 & 355M & 7.34 & 152.8 \\
		ARM & \scriptsize \textbf{DQ-Transformer$_l$} & 640 & 655M & \textbf{5.11} & \textbf{178.2} \\
		\bottomrule
	\end{tabular}
	\caption{\footnotesize Comparison of class-conditional generation with million-level parameters on ImageNet. $L$ is coding length. ARM denotes for autoregressive model. * denotes for our reproduction.}
	\label{class_million}
\end{table}

\textbf{Unconditional generation.} As shown in Table \ref{unconditional_autoregression}, our model outperform all existing autoregressive state-of-the-art models including the strongest large-scale ViT-VQGAN\cite{yu2021vector} by a 7.4\% quality improvement. We compare with other types of state-of-the-art models in Table \ref{unconditional_other} and also achieve top-level performance. The qualitative results of unconditional generation are shown on the left of Figure \ref{visual}. 

\textbf{Class-conditional generation.} The comparison is split into Million/Billion according to whether they can be trained under normal computing resources (\ie, 24G memory 3090). We first compare with all million-level parameters state-of-the-art in Table \ref{class_million}. Our model with $355$M parameters already outperforms all autoregressive and diffusion models. Moreover, our model with $655$M outperforms GAN-based and bi-direct state-of-the-art, which demonstrates our effectiveness and scalability. We further compare with large-scale billion-level state-of-the-art in Table \ref{class_billion}, where we achieve top-level performance with fewer parameters. The qualitative results of class-conditional generation are shown on the right of Figure \ref{visual}. 


\begin{table}[t]
    \footnotesize
	\centering
	\begin{tabular}{l c c c c c}
		\toprule
		Type & Model & $L$ & \#Params & FID$\downarrow$ & IS$\uparrow$ \\
		\hline
		Diffusion & ImageBART\cite{esser2021imagebart} & - & 3.5B & 21.19 & 61.6 \\
		ARM & VQ-VAE-2\cite{razavi2019generating} & 5120& 13.5B & 31.11 & ~45 \\
		ARM & VQGAN\cite{esser2021taming} & 256 &  1.4B & 15.78 & 78.3 \\
		ARM & ViT-VQGAN\cite{tang2022improved} & 1024 & 2.2B & 4.17 & 175.1 \\
		ARM & RQ-VAE\cite{lee2022autoregressive} & 256 & 3.8B & 7.55 & 134 \\
		\hline
		ARM &  DQ-Transformer$_b$ & 640 & 355M & 7.34 & 152.8 \\
		ARM & \textbf{DQ-Transformer$_l$} & 640 & \textbf{655}M & \textbf{5.11} & \textbf{178.2} \\
		\bottomrule
	\end{tabular}
	\caption{\footnotesize Comparison between \textbf{our million-level model} and \textbf{large-scale billion-level big models} of class-conditional generation on ImageNet.}
	\label{class_billion}
\end{table}

\begin{table}[t]
    \footnotesize
	\centering
	\begin{tabular}{l c c}
		\toprule
		Model Type & Model & FID$\downarrow$ \\
		\hline
		GAN & BigGAN\cite{brock2018large} & 12.4 \\
		GAN & StyleGAN2\cite{karras2020analyzing} & 3.8 \\
		VAE & VDVAE\cite{child2020very} & 28.5 \\
		Diffusion & ImageBART\cite{esser2021imagebart} & 9.57 \\
		Diffusion & UDM\cite{kim2021score} & 5.54 \\
		\hline
		\textbf{Autoregressive} & \textbf{DQ-Transformer$_b$} & \textbf{4.91} \\
		\bottomrule
	\end{tabular}
	\caption{\footnotesize Comparison with other types of state-of-the-art on unconditional FFHQ, where we further improve the quality of autoregressive models.}
	\label{unconditional_other}
\end{table}


\subsection{Ablations $\&$ Analysis}

\begin{figure*}
  \centering
  \includegraphics[width=1.0\linewidth]{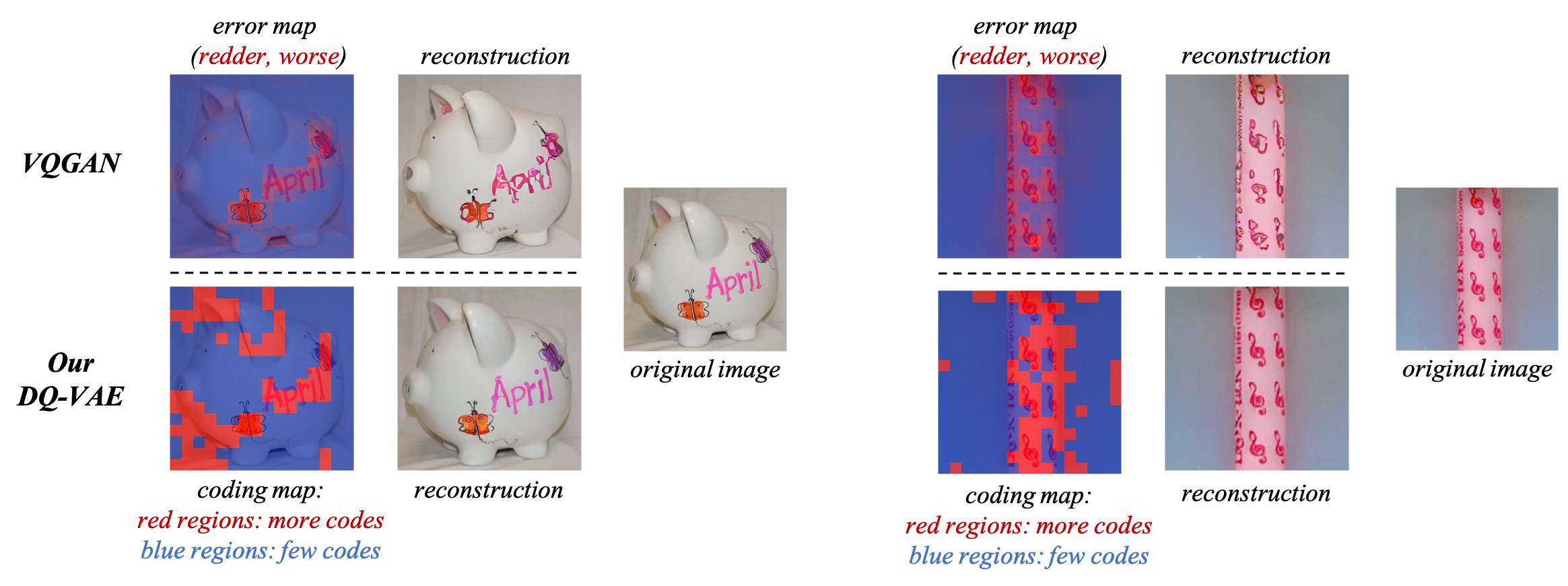}
  \caption{Visualization of the variable-length coding of our DQ-VAE, where \emph{\textbf{our coding map exactly matches the error map of VQGAN}} and therefore leads to better reconstruction quality, \ie, the \textcolor{red}{\emph{\textbf{information-dense regions}}} where VQGAN has \emph{\textbf{higher reconstruction error}} are assigned to \textcolor{red}{\emph{\textbf{more codes}}}, while \textcolor{blue}{\emph{\textbf{information-sparse regions}}} where VQGAN has \emph{\textbf{lower reconstruction error}} are assigned to \textcolor{blue}{\emph{\textbf{few codes}}}.}
  \label{visual_assignment}
\end{figure*}

\textbf{Analysis on DQ-VAE.} We first demonstrate that our variable-length coding has better reconstruction compared to the existing fixed-length one in Table \ref{ablation_1}. We take VQGAN\cite{esser2021taming} of $f=16$ as the baseline, and DQ-VAE adopts triple granularities of $\rm F = \{8,16,32\}$ and subject to:

\begin{equation}
    r_{f=32} = 4 \times r_{f=8},
\end{equation}
which ensures DQ-VAE's expected mean code length is the same as VQGAN (\ie, 256). We could conclude: (1) With a proper ratio, DQ-VAE's variable-length coding achieves better reconstruction quality compared to VQGAN's fixed-length one (ours 4.08 \vs VQGAN's 4.82). The reason is that important regions require more codes to encode necessary information, while fewer codes are enough for unimportant ones since they are less informative. The phenomenon also reveals that existing fixed-length coding is both insufficient in important regions and redundant in unimportant ones. (2) When we improperly increase $r_{f=8}$, we get a larger $r_{f=32}$ which will inevitably assign some important regions with fewer codes and thus degrade the reconstruction quality. (3) Moreover, DQ-VAE's adaptive assignment significantly outperforms the random one (ours 4.08 \vs random's 7.32) which demonstrates that DQ-VAE could distinguish important regions from unimportant ones.

\begin{table}[t]
    \footnotesize
	\centering
	\begin{tabular}{l c c c}
		\toprule
		Model & $\rm F$ & ratio & rFID$\downarrow$\\
		\hline
		VQGAN\cite{esser2021taming} & 16 & - & 4.82 \\
        DQ-VAE & \{8,16,32\} & \{0.05, 0.75, 0.3\} & 4.57  \\
		DQ-VAE & \{8,16,32\} & \{0.075, 0.625, 0.3\} & 4.08  \\
		DQ-VAE / random & \{8,16,32\} & \{0.075, 0.625, 0.3\} & 7.32  \\
		DQ-VAE & \{8,16,32\} & \{0.1, 0.5, 0.4\} & 4.96 \\
		DQ-VAE & \{8,16,32\} & \{0.125, 0.375, 0.5\} & 6.39 \\
		\bottomrule
	\end{tabular}
	\caption{\footnotesize Ablations of the proposed variable-length coding on ImageNet. Here $\rm F$ denotes the granularity candidates set. ``ratio" denotes the ratio of each granularity. We show that variable-length coding could bring better reconstruction compared to fixed-length coding on the same code length.}
	\label{ablation_1}
\end{table}

\begin{table}[t]
    \footnotesize
	\centering
	\begin{tabular}{c c c c c c}
		\toprule
		Model & $r_{f=8}$ & mean (expected) & var & rFID$\downarrow$ & usage $\uparrow$\\
		\hline
		VQGAN & 0 & 256 & - & 4.46 & 63.89\% \\
		DQ-VAE & 0.1 & 332 (333) & 760.6 & 3.6 & 63.02\% \\
		DQ-VAE & 0.3 & 494 (486) & 621.3 & 3.02 & 62.3\% \\
		DQ-VAE & 0.5 & 646 (640) & 348.3 & 2.38 & 59.9\% \\
		DQ-VAE & 0.7 & 792 (794) & 285.6 & 2.09 & 56.01\% \\
		DQ-VAE & 0.9 & 945 (947) & 132.4 & 1.87 & 52.32\% \\
		VQGAN & 1 & 1024 & - & 1.8 & 46.41\% \\
		\bottomrule 
	\end{tabular}
	\caption{\footnotesize Ablations of different granularity ratios of DQ-VAE with $\rm F$=\{8, 16\} on FFHQ. Here $r_{f=8}$ denotes the ratio of $f=8$. ``mean'' and ``var'' denote the mean and variance of dynamic coding length. The codebook usage is calculated as the percentage of used codes over the entire test set.}
	\label{ablation_2}
\end{table}

We then analyze the impact of different ratio percentages in Table \ref{ablation_2}, where DQ-VAE adopts dual granularities of $\rm F=\{8,16\}$. We show that: (1) The mean code length of each ratio matches the expectation well, which validates our proposed budget loss. (2) The results are consistent with \emph{the Pareto principle}, which is also known as \emph{20/80 laws}. To be specific, when increasing $r_{f=8}$ from 0 to 0.3, we get 1.44 FID improvement while only a slight codebook usage drop, which indicates that the first 30\% percentage important regions contribute the most valid information of images and existing fixed-length coding is insufficient in them. Meanwhile, when increasing $r_{f=8}$ from 0.7 to 1.0, we only get a subtle 0.21 FID improvement but a significant 9.6\% codebook usage drop, which indicates that the last 30\% percentage unimportant regions contribute little valid information of images but most redundancy. The experimental results strongly support our motivations for variable-length coding to get rid of insufficiency and redundancy simultaneously.

We visualize our variable-length coding on ImageNet in Figure \ref{visual_assignment}, where DQ-VAE adopts dual granularities of $\rm F=\{8,16\}$ and $r_{f=8}=0.3$. The error map is calculated by $l_1$ loss of each $16^2$ size region between images and VQGAN ($f=16$) reconstructions. The red regions in our coding map are assigned to $f=8$ (4 codes) while the blue ones are assigned to $f=16$ (1 code). We show that our coding map matches VQGAN's error map, \ie, important regions are assigned to more codes and unimportant ones are assigned to few codes, leading to better reconstruction quality. 

\textbf{Analysis on the effectiveness of DQ-Transformer}. We first validate our input layers designs in Table \ref{ablation_3}. The \emph{non-shared-position} and \emph{granularity} layers are very important since they distinguish different granularities. Without these designs, DQ-Transformer fails to know which granularity of code should
be generated next, and thus performs worse.

\begin{table}[t]
    \footnotesize
 \centering
 \begin{tabular}{c c c c c}
  \toprule
  Content & Position & Granularity & Absolute position & FID$\downarrow$\\
  \hline
  shared & non-shared & \color{blue}\Checkmark & \color{blue}\Checkmark & 4.91 \\
  non-shared & non-shared & \color{blue}\Checkmark & \color{blue}\Checkmark & 5.54 \\
  shared & shared & \color{blue}\Checkmark & \color{blue}\Checkmark & 18.28 \\
  shared & non-shared & \color{red}\XSolidBrush & \color{blue}\Checkmark & 16.87 \\
  shared & non-shared & \color{blue}\Checkmark & \color{red}\XSolidBrush & 5.06 \\
  \bottomrule
 \end{tabular}
 \caption{\footnotesize Ablations of DQ-Transformer input designs on FFHQ. Here ``granularity" denotes for DQ-Transformer’s granularity layer.}
 \label{ablation_3}
\end{table}

\begin{figure}
  \centering
  \includegraphics[width=1.0\linewidth]{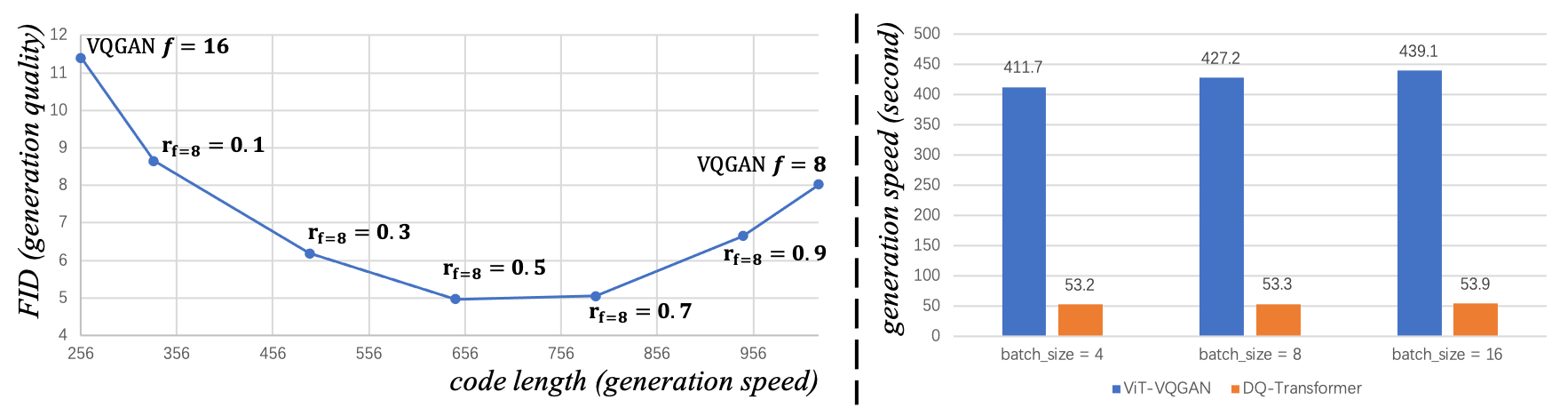}
  \caption{\footnotesize Left: The Pareto curves of the different ratios between generation quality (FID) and generation speed (code length) on FFHQ. Right: The speed comparison between large-scale ViT-VQGAN\cite{yu2021vector} and our DQ-Transformer(base) according to different batch sizes on FFHQ.}
  \label{pareto}
\end{figure}

We then analyze the generation quality of different ratios in Figure \ref{pareto} left. The generation speed of autoregressive models mostly depends on their code length. The Pareto curve shows that the generation quality (FID) saturates when $r_{f=8}$ reaches 0.5. The experimental phenomenon reveals that a proper ratio is important for the unity of a high generation quality and fast generation speed since it guarantees effective coding in both important regions and unimportant ones for an accurate $\&$ compact code representation. 

\textbf{Analysis on the efficiency of DQ-Transformer}. We compare our generation speed to the existing state-of-the-art autoregressive model ViT-VQGAN\cite{yu2021vector} according to different batch sizes in Figure \ref{pareto} right. The generation speeds are evaluated on a single RTX-3090 GPU and the setup of ViT-VQGAN is implemented the same as its original paper. Our model achieves a much faster generation speed for all batch sizes which validates the efficiency brought by our accurate and compact code representation. 


\section{Conclusion \& Future Direction}

In this study, we point out that the existing fixed-length coding ignores the naturally different information densities of image regions and is inherently limited by insufficiency and redundancy, which degrades generation quality and speed. Moreover, the fixed-length coding brings an unnatural raster-scan autoregression. We thereby propose a novel two-stage generation framework: (1) \emph{DQ-VAE} which dynamically assigns variable-length codes to regions based on their information densities for an accurate and compact code representation. (2) \emph{DQ-Transformer} which then models the position and content of codes alternately, generating images autoregressively in a more natural and effective coarse-to-fine order for the first time. To effectively teach the difference between different granularities, we further design \emph{shared-content}, \emph{non-shared-position}, and \emph{granularity} input layers. Comprehensive experiments on various image generations validate our effectiveness and efficiency.

\textbf{Future Direction.} VQ is the foundation for modern autoregressive\cite{ramesh2021zero, lee2022autoregressive, zheng2022movq, ding2021cogview, yu2022scaling}, discrete diffusion\cite{rombach2022high, gu2022vector}, and bidirectional\cite{chang2022maskgit} generation, and even pretraining\cite{bao2021beit,mao2021discrete}. Our study validates the effectiveness and efficiency of the variable-length coding for autoregressive generation, but its great potential for diffusion, bi-direction, and pretraining is worth further exploration in the future.

\section{Acknowledgments}
This work is supported by National Natural Science
Foundation of China under Grant 62222212 and Science Fund for Creative Research Groups under Grant 62121002.

{\small
\bibliographystyle{ieee_fullname}
\bibliography{main}
}

\end{document}